\documentclass[10pt,twocolumn,letterpaper]{article}

\usepackage{times}
\usepackage{epsfig}
\usepackage{graphicx}
\usepackage{amsmath}
\usepackage{amssymb}
\usepackage{algorithmic}
\usepackage[linesnumbered,ruled,vlined]{algorithm2e}
\usepackage{multirow}
\usepackage{booktabs}
\usepackage{siunitx}
\usepackage{bm}
\usepackage{amsfonts}

\usepackage{CJK}
\newif\ifdraft\drafttrue

\ifdraft

\newcommand\xw[1]{{\footnotesize \color{red}[#1 - \textbf{Xiaofang}]}}

\else
\newcommand\todo[1]{}
\newcommand\xw[1]{}
\newcommand\mpet[1]{}
\fi

\usepackage[pagebackref=true,breaklinks=true,letterpaper=true,colorlinks,bookmarks=false]{hyperref}

\begin{document}
\begin{CJK*}{UTF8}{gkai}
\title{Close Yet Discriminative Domain Adaptation}
\author{Lingkun Luo$^{1\ast}$,   Xiaofang Wang $^{2}$ \thanks{These first two authors contributed equally.},   Shiqiang Hu $^{1}$, Chao Wang $^{1}$, Yuxing Tang $^{2}$, Liming Chen $^{2}$  \\
$^1 $ School of Aeronautics and Astronautics, Shanghai Jiao Tong University, Shanghai, China. \\ $\{lolinkun1988, sqhu, wang\_chao\}$@sjtu.edu.cn \\
$^2$  LIRIS, CNRS UMR 5205, \'Ecole Centrale de Lyon, 36 avenue Guy de Collongue, \'Ecully, F-69134, France. \\ $\{xiaofang.wang, yuxing.tang, liming.chen\}$@ec-lyon.fr
}
%

\maketitle


\begin{abstract}
Domain adaptation is transfer learning which aims to generalize a learning model across training and testing data with different distributions. Most previous research tackle this problem in seeking a shared feature representation between source and target domains while reducing the mismatch of their data distributions. In this paper, we propose a close yet discriminative domain adaptation method, namely CDDA, which generates a latent feature representation with two \emph{interesting} properties. First, the discrepancy between the source and target domain, measured in terms of both  marginal and conditional probability distribution via Maximum Mean Discrepancy is minimized so as to \emph{attract} two domains close to each other. More importantly, we also design a repulsive force term, which maximizes the distances between each label dependent sub-domain to all others so as to \emph{drag} different class dependent sub-domains far away from each other and thereby increase the discriminative power of the  adapted domain. Moreover, given the fact that the underlying data manifold could have complex geometric structure, we further propose the constraints of label smoothness and geometric structure consistency for label propagation. Extensive experiments are conducted on  36 cross-domain image classification tasks over four public datasets. The Comprehensive results show that the proposed method consistently outperforms the state-of-the-art methods with significant margins.
\end{abstract}
\section{Introduction}

Thanks to deep networks, recent years have witnessed impressive progress in an increasing number of machine learning and computer vision tasks, \textit{e.g.},  image classification\cite{ILSVRC15,He2015}, object detection \cite{Everingham15,girshickICCV15fastrcnn}, semantic segmentation \cite{Cordts2016Cityscapes,Everingham15,zhao2016pspnet}. However, these impressive progress have been made possible only when massive amount of labeled training data are available and such a requirement hampers their adoption to a number of real-life applications where labeled training data don't exist or not enough in quantity. On the other hand, manual annotation of large training data could be extremely tedious and prohibitive for a given application. An interesting solution to this problem is transfer learning through \textit{domain adaptation} \cite{pan2010survey}), which aims to leverage abundant existing labeled data from a different but related domain (source domain) and generalize a predictive model learned from the source domain to unlabeled target data (target domain) despite the discrepancy between the source and target data distributions.       

\begin{figure*}[h!]
	\centering
	\label{fig:diff}
	\includegraphics[width=1\linewidth]{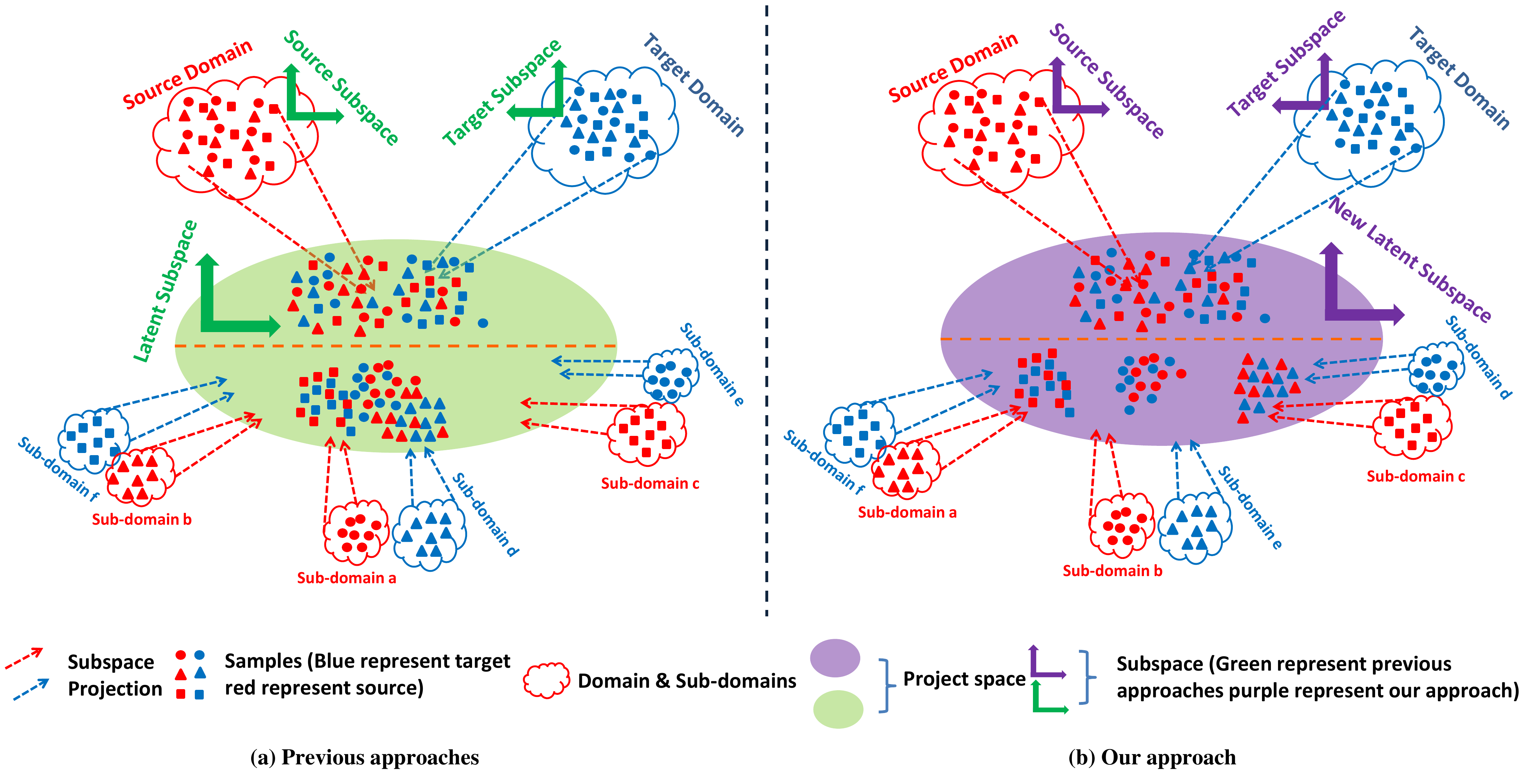}
	\vspace{-7pt}
	\caption {  Illustration of the major difference between our proposed method and previous state-of-the-art: The geometrical shape in round, triangle and square represents samples of  different class labels. Cloud colored in red or blue represents the source or target domain, respectively. The latent shared feature space is represented by ellipse.  The green ellipse illustrates the the latent feature space obtained by the previous approaches, whereas the purple one illustrates the novel latent shared feature space by the proposed method. The upper part of both ellipses represents the marginal distribution,  while the lower part denotes the conditional distribution. As can be seen from the marginal distribution in the lower part of Fig.1(b), samples with same label are clustered together  while samples with different labels, thus from different sub-domains, are separated. This is in contrast with the conditional distribution in the lower part of Fig.1(a) where samples with different labels are completely mixed, thus making harder the discrimination of samples of different labels.} 
\end{figure*}

The core idea of most proposed methods for domain adaptation is to reduce the discrepancy between domains and learn a domain-invariant predictive model from data. State of the art has so far featured two  mainstream algorithms in reducing data distribution discrepancy: (1) feature representation transfer, which aims to find "good" feature representations to minimize domain differences and the error of classification or regression models;  and (2) instance transfer, which attempts to re-weight some "good" data from source domain, which may be useful for the target domain.  It minimizes the distribution differences by re-weighting the source domain data and then trains a predictive model on the re-weighted source data. 


In this paper, we are interested in feature representation transfer which seeks a domain invariant latent space, while preserving at the same time important structure of original data, \textit{e.g.}, data variance or geometry. Early methods, \textit{e.g.}, \cite{blitzer2006domain}, propose a structural correspondence learning (SCL), which first defines a set of pivot features and then identifies correspondences among features from different domains by modeling their correlations with the pivot features.  Later, transfer learning problems are approached via dimensionality reduction.  \cite{pan2011domain} learns a novel feature representation across domains in a Reproducing Kernel Hilbert Space  with the Maximum Mean Discrepancy (MMD) measure \cite{borgwardt2006integrating}, through the so-called transfer component analysis (TCA). TCA \cite{pan2011domain} is an extension of \cite{pan2008transfer}, with the purpose to reduce computational burden.  \cite{long2013transfer} goes one step further and remarks that both marginal and conditional distribution could be different between the source and target domains. As a result,  Joint Distribution Adaptation (JDA) is proposed to jointly minimize the mismatches of marginal and conditional probability  distributions. The previous research has thus so far only focused on matching marginal and/or conditional distributions for transfer learning while ignoring the discriminative properties to be reinforced between different classes in the adapted domain.  

In this paper, we propose to extract a latent shared feature space underlying the domains where the discrepancy between domains is reduced but more importantly, the original discriminative information between classes is simultaneously reinforced. Specifically, not only we seek to find a shared feature space in minimizing the discrepancy of both marginal and conditional probability distributions as in JDA \cite{long2013transfer},  but also introduce a discriminative model, called subsequently as \textit{repulsive force}, in light of the  Fisher’s linear discriminant analysis (FLDA) \cite{fisher1936use}. This repulsive force \emph{drags} the sub-domains with different labels far away from each other in maximizing their distances measured in terms of \textit{Maximum Mean Discrepancy} (MMD), thereby making more discriminative  data from different sub-domains. This is in clear contrast to the previous approaches as illustrated in Fig.\ref{fig:diff}. Most previous works, \textit{e.g.},JDA, only seek to align marginal or conditional distributions between the source and target domain and the resultant latent subspace therefore falls short in terms of discrimination power as illustrated in the lower part of the green ellipse of Fig.\ref{fig:diff}(a), where samples of different labels are all mixed up.  In contrast, as can be seen in the lower part of the purple ellipse of Fig.\ref{fig:diff}(b), the proposed method unifies the decrease of data distribution discrepancy and the increase of the discriminative property between classes into a same framework and finds a novel latent subspace where samples with same label are put close to each other while samples with different labels are well separated. Moreover, given the fact that the manifold of both source and target data in the shared latent feature space could have complex geometric structure, we further propose label propagation based on the respect of two constraints, namely label smoothness consistency (LSC) and geometric structure consistency (GSC), for the prediction of target data labels. That is, a good label propagation should well preserve the label information(constraint LSC) and not change too much from the shared data manifold (constraint GSC).

To sum up, the contributions in this paper are threefold: 

\begin{itemize}
	\item A novel repulsive force is proposed to increase the discriminative power of the shared latent subspace, aside of decreasing both the marginal and conditional distributions between the source and target domains. 
    
	\item Unlike a number of domain adaptation methods, \textit{e.g.}, JDA \cite{long2013transfer}, which use Nearest Neighbor(NN) with Euclidean distance to predict labels in target domain,  the prediction in the proposed model, is deduced via label propagation in respect of the underlying data manifold geometric structure. 
%
%
    
	\item Extensive experiments are conducted on comprehensive datasets, and verify the effectiveness of the proposed method which outperforms  state-of-the-art domain adaptation algorithms with a significant margin.  
\end{itemize}

    
    

The rest of the paper is organized as follows. In Section 2, we discuss previous works related to ours and highlight their differences. In Section 3, first we describe the problem and preliminaries of domain adaptation and then we present our proposed method. Experiment results and discussions are presented in Section 4 and finally we draw the conclusion in Section 5.

\vspace{-5pt}

\section{Related Work}

In this section, we discuss previous works which are related to our method and analyze their differences. 

In machine learning,  domain adaptation is transfer learning which aims to learn an effective predictive  model for a target domain without labeled data in leveraging abundant existing labeled data of a different but related source domain.   Because the collection of large labeled data as needed in traditional machine learning is often prohibitive for many real-life applications, there is an increasing interest on this \emph{young} yet \emph{hot} topic \cite{pan2010survey}\cite{weiss2016survey}.    According to the taxonomy made in recent surveys \cite{pan2010survey}\cite{weiss2016survey} \cite{long2013transfer}, the proposed method falls down into the feature representation category.


Recent popular methods embrace the dimensionality reduction  to seek a latent shared feature space between the source and the target domain. Its core idea is to project the original data into a low-dimensional latent space  with preserving important structure of original data. However,  \cite{pan2008transfer} points out that direct application of Principal Component Analysis (PCA) can not guarantee the preservation of  discriminative data structures. Their proposed remedy is to maximize the variance of the embedded data. Another interesting idea in  \cite{pan2008transfer} is the use of a nonparametric criterion, namely  \textit{Maximum Mean Discrepancy} (MMD),  based on Reproducing Hilbert Space (RKHS) \cite{borgwardt2006integrating}, to estimate the distance between two distributions. Later, \cite{pan2011domain} further improves \cite{pan2008transfer}  in terms of  computational efficiency. With JDA,  \cite{long2013transfer} goes one step further and propose not only to minimize the mismatch of the  cross-domains marginal probability distributions but also their conditional probability distributions  based on the framework of \cite{pan2008transfer,pan2011domain}. The proposed framework in this paper can be considered as an extension of JDA with two major differences. First, we seek not only for a latent subspace which minimizes the mismatch of both the marginal and conditional probability distributions across domains, but also reinforces the discriminative structure of sub-domains in original data. We achieve this goal in introducing a novel term which acts as repulsive force to drag away different sub-domains both in source and target domain, respectively. 

Note that we do not discuss the line of work in the literature on transfer learning which is embedded into deep convolutional neural network as the features used in this work are not deep features; Nevertheless we have noticed their impressive performance, thanks to the combination of the  latest advances in transfer learning discussed above with the cutting-edge  understanding on the transferability  \cite{glorot2011domain} of state-of-the-art deep neural networks, \textit{e.g.},  Deep Adaptation Network(DAN) \cite{long2015learning},  \textit{etc.} Mixing seamlessly our proposed transfer knowledge model with state-of-the-art deep networks will be the subject of our upcoming investigation. 
\vspace{-5pt}

\section{Close Yet Discriminative Domain Adaptation}
In this section, we present in detail the proposed Close yet Discriminative Domain Adaptation (CDDA) method.

\subsection{Problem Statement}
We begin with the definitions of notations and concepts most of which we borrow directly  from \cite{long2013transfer}. 

A domain $D$ is defined as an m-dimensional feature space $\chi$ and a marginal probability distribution $P(x)$, \textit{i.e.}, $\mathcal{D}=\{\chi,P(x)\}$ with $x\in \chi$. 

Given a specific domain $D$, a  task $T$ is composed of a C-cardinality label set $\mathcal{Y}$  and a classifier $f(x)$,\textit{ i.e.}, $T = \{\mathcal{Y},f(x)\}$, where $f({x}) = \mathcal{Q}( y |x)$ which can be interpreted as the class conditional probability distribution for each input sample $x$. 
%
%

In unsupervised domain adaptation, we are given a source domain $\mathcal{D_S}=\{x_{i}^{s},y_{i}^{s}\}_{i=1}^{n_s}$ with $n_s$ labeled samples, and  a unlabeled target domain $\mathcal{D_T}=\{x_{j}^{t}\}_{j=1}^{n_t}$ with $n_t$  unlabeled samples with the assumption that source domain $\mathcal{D_S}$ and target domain $\mathcal{D_T}$ are different, \textit{i.e.},  $\mathcal{\chi}_S=\mathcal{{\chi}_T}$, $\mathcal{Y_S}=\mathcal{Y_T}$, $\mathcal{P}(\mathcal{\chi_S}) \neq \mathcal{P}(\mathcal{\chi_T})$, $\mathcal{Q}(\mathcal{Y_S}|\mathcal{\chi_{S}}) \neq \mathcal{Q}(\mathcal{Y_T}|\mathcal{\chi_{T}})$.
We also define the notion of sub-domain, denoted as ${\cal D}_{\cal S}^{(c)}$, representing the set of samples in ${{\cal D}_{\cal S}}$ with label $c$. Similarly, a sub-domain ${\cal D}_{\cal T}^{(c)}$ can be defined for the target domain as the set of samples in ${{\cal D}_{\cal T}}$ with label $c$. However, as ${{\cal D}_{\cal T}}$is the target domain with unlabeled samples, a basic classifier,\textit{ e.g.}, NN, is needed to attribute  pseudo labels for samples in ${{\cal D}_{\cal T}}$.

The aim of the Close yet Discriminative Domain Adaptation (CDDA) is to learn a latent feature space with following properties: 1) the distances of both marginal and conditional probability of source and target domains are reduced; 2)  The distances between each sub-domain to the others,  are increased in order to  push them far away from each other; 3) The deduction of  label prediction is imposed via  two constraints, \textit{i.e.},  label consistency and geometric structure of label space.

\vspace{-3pt}
\subsection{Latent Feature Space with Dimensionality Reduction}
The finding of a latent feature space  with dimensionality reduction  has been demonstrated useful in several previous works, \textit{e.g.}, \cite{pan2008transfer,pan2011domain,long2013transfer}, for domain adaptation.  One of its important properties is that  original data is projected to a lower dimensional space which is considered as \emph{principal} structure of data. In the proposed method, we also apply the Principal Component Analysis (PCA).  Mathematically, given with an input data matrix $\boldsymbol{X} = [{\mathcal{D_S}},\mathcal{D_T}]$, $\boldsymbol{X} \in {\mathbb{R}^{m\times({n_s} + {n_t})}}$, the centering matrix is defined as  $\boldsymbol{H} = \boldsymbol{I} - \frac{1}{n_s+n_t}\boldsymbol{1}$, where $\boldsymbol{1}$ is the $(n_s+n_t) \times (n_s+n_t)$ matrix of ones. The optimization of PCA is to find a projection space $\boldsymbol{A}$ which  maximizes the embedded data variance.
\begin{equation}\label{eq:pca}
	\begin{array}{c}
		\mathop {\max}\limits_{\boldsymbol{A^TA} = \boldsymbol{I}} tr(\boldsymbol{A}^T\boldsymbol{ XH}\boldsymbol{X}^T \boldsymbol{A})
	\end{array}
\end{equation}
where $tr(\mathord{\cdot})$ denotes the trace of a matrix,   $\boldsymbol{XH}\boldsymbol{X}^T$ is the data covariance matrix, and $\bf A \in \mathbb{R}^{m \times k}$ with $m$ the feature dimension and $k$ the dimension of the projected subspace. The optimal solution  is calculated by solving an eigendecomposition problem: $\boldsymbol{XH}\boldsymbol{X}^T=\boldsymbol{A\Phi}$, where $\boldsymbol{\Phi}=diag(\phi_1,\dots, \phi_k )$ are the $k$ largest eigenvalues. Finally, the original data $\boldsymbol{X}$ is projected into the  optimal $k$-dimensional subspace using $\boldsymbol{Z} = \boldsymbol{A}^T\boldsymbol{X}$.

\subsection{ Closer: Marginal and Conditional Distribution Domain Adaptation}
However, the feature space calculated via PCA is not sufficiently \emph{good} enough for our problem of domain adaptation problem, for PCA only seeks to maximize the variance of the projected data from the two domains and does not explicitly reduce their distribution mismatch \cite{long2013transfer,long2015learning}.     Since the distance of data distributions across domain  can also be empirically measured , we explicitly leverage the nonparametric distance measurement MMD in RKHS \cite{borgwardt2006integrating} to compute the distance between expectations of source domain and target domain, once the original data projected into  a low-dimensional feature space via.  Formally, the empirical distance of the two domains  is defined as:
\begin{equation}\label{eq:marginal}
	\begin{array}{l}		
		Dis{t^{marginal}}({{\cal D}_{\cal S}},{{\cal D}_{\cal T}}) =\\ {\left\| {\frac{1}{{{n_s}}}\sum\limits_{i = 1}^{{n_s}} {{{\bf{A}}^T}{x_i} - } \frac{1}{{{n_t}}}\sum\limits_{j = {n_s} + 1}^{{n_s} + {n_t}} {{{\bf{A}}^T}{x_j}} } \right\|^2}
		= tr({{\bf{A}}^T}\bf{X}{\bf{M_0}}\bf{{X^T}A})		
		\end{array}
\end{equation}

where ${{\bf{M}}_0}$ represents the marginal distribution between ${{\cal D}_{\cal S}}$ and ${{\cal D}_{\cal T}}$ and its calculation is obtained by:
	
	\begin{equation}\label{eq:M0}
	\begin{array}{l}
	{({{\bf{M}}_0})_{ij}} = \left\{ \begin{array}{l}
	\frac{1}{{{n_s}{n_s}}},\;\;\;{x_i},{x_j} \in {D_{\cal S}}\\
	\frac{1}{{{n_t}{n_t}}},\;\;\;{x_i},{x_j} \in {D_{\cal T}}\\
	0,\;\;\;\;\;\;\;\;\;\;\;\;otherwise
	\end{array} \right.
	\end{array}
	\end{equation}
    

where ${x_i},{x_j} \in (\mathcal{D_S} \cup \mathcal{D_T})$. The difference between the marginal distributions $\mathcal{P}(\mathcal{X_S})$ and $\mathcal{P}(\mathcal{X_T})$ is reduced in minimizing {$Dis{t^{marginal}}({{\cal D}_{\cal S}},{{\cal D}_{\cal T}})$}.

Similarly, the distance of conditional probability distributions is defined as the sum of the empirical distances over the class labels between the sub-domains of a same label in the source and target domain: 
\begin{equation}\label{eq:conditional}
		\begin{array}{c}
		\begin{array}{l}
		Dis{t^{conditional}}\sum\limits_{c = 1}^C {({{\cal D}_{\cal S}}^c,{{\cal D}_{\cal T}}^c)}  = \\
		{\left\| {\frac{1}{{n_s^{(c)}}}\sum\limits_{{x_i} \in {{\cal D}_{\cal S}}^{(c)}} {{{\bf{A}}^T}{x_i}}  - \frac{1}{{n_t^{(c)}}}\sum\limits_{{x_j} \in {{\cal D}_{\cal T}}^{(c)}} {{{\bf{A}}^T}{x_j}} } \right\|^2}\\
		= tr({{\bf{A}}^T}{\bf{X}}{{\bf{M}}_c}{{\bf{X}}^{\bf{T}}}{\bf{A}})
		\end{array}
		\end{array}
\end{equation}
where $C$ is the number of classes, $\mathcal{D_S}^{(c)} = \{ {x_i}:{x_i} \in \mathcal{D_S} \wedge y({x_i} = c)\} $ represents the ${c^{th}}$ sub-domain in the source domain, $n_s^{(c)} = {\left\| {\mathcal{D_S}^{(c)}} \right\|_0}$ is the number of samples in the ${c^{th}}$ {source} sub-domain. $\mathcal{D_T}^{(c)}$ and $n_t^{(c)}$ are defined similarly for the target domain. Finally, $\bf M_c$ represents the conditional distribution between sub-domains in ${{\cal D}_{\cal S}}$ and ${{\cal D}_{\cal T}}$ and it is defined as: 
\begin{equation}\label{eq:marginal}
\begin{array}{*{20}{c}}
{{{({{\bf{M}}_c})}_{ij}} = \left\{ {\begin{array}{*{20}{l}}
		{\frac{1}{{n_s^{(c)}n_s^{(c)}}},\;\;\;{x_i},{x_j} \in {D_{\cal S}}^{(c)}}\\
		{\frac{1}{{n_t^{(c)}n_t^{(c)}}},\;\;\;{x_i},{x_j} \in {D_{\cal T}}^{(c)}}\\
		{\frac{{ - 1}}{{n_s^{(c)}n_t^{(c)}}},\;\;\;\left\{ {\begin{array}{*{20}{l}}
				{{x_i} \in {D_{\cal S}}^{(c)},{x_j} \in {D_{\cal T}}^{(c)}}\\
				{{x_i} \in {D_{\cal T}}^{(c)},{x_j} \in {D_{\cal S}}^{(c)}}
				\end{array}} \right.}\\
		{0,\;\;\;\;\;\;\;\;\;\;\;\;otherwise}
		\end{array}} \right.}
\end{array}
\end{equation}


In minimizing ${Dis{t^{conditional}}\sum\limits_{c = 1}^C {({D_{\cal S}}^c,{D_{\cal T}}^c)} }$,  the mismatch of conditional distributions between ${{D_{\cal S}}^c}$ and ${{D_{\cal T}}^c}$ is reduced.

\subsection{More discriminative:Repulsive Force Domain Adaptation}

The latent feature subspace obtained by the joint marginal and conditional domain adaptation as in JDA, is to reduce the differences between the source and target domain. As such, two spaces of data are \emph{attracted}  to be close to each other. However, their model has ignored an important property for the elaboration of an effective predictor,  \textit{i.e.},  the preservation or reinforcement of  discriminative information related to sub-domains. In this paper, we introduce a novel \textit{repulsive force} domain adaption, which aims to increase the distances of sub-domains with different labels, so as to improve the discriminative power of the latent shared features and thereby making it possible better predictive model for the target domain. To sum up, we aim to generate a latent feature space where the discrepancy between domains is reduced while simultaneously the distances between sub-domains of different labels are increased for an reinforced discriminative power of the underlying latent feature space.

Specifically, the repulsive force domain adaptation is defined as: 
$Dis{t^{repulsive}} = Dist_{{\cal S} \to {\cal T}}^{repulsive} + Dist_{{\cal T} \to {\cal S}}^{repulsive}$, where ${{\cal S} \to {\cal T}}$ and ${{\cal T} \to {\cal S}}$ index the distances computed from ${D_{\cal S}}$ to ${D_{\cal T}}$ and ${D_{\cal T}}$ to ${D_{\cal S}}$, respectively. $Dist_{{\cal S} \to {\cal T}}^{repulsive}$ represents the sum of the distances between each source sub-domain ${D_{\cal S}}^{(c)}$ and all the  target sub-domains ${D_{\cal T}}^{(r);\;r \in \{ \{ 1...C\}  - \{ c\} \} }$ except the one with the label $c$. The sum of these distances is explicitly defined as:
	\begin{equation}\label{eq:StoT}
		\resizebox{1\hsize}{!}{%
			${Dist}_{{\cal S} \to {\cal T}}^{repulsive} = \sum\limits_{c = 1}^C \begin{array}{l}
			{\left\| {\frac{1}{{n_s^{(c)}}}\sum\limits_{{x_i} \in {D_{\cal S}}^{(c)}} {{{\bf{A}}^T}{x_i}}  - \frac{1}{{\sum\limits_{r \in \{ \{ 1...C\}  - \{ c\} \} } {n_t^{(r)}} }}\sum\limits_{{x_j} \in D_{\cal T}^{(r)}} {{{\bf{A}}^T}{x_j}} } \right\|^2}\\
			= \sum\limits_{c = 1}^C {tr({{\bf{A}}^T}{\bf{X}}{{\bf{M}}_{{\cal S} \to {\cal T}}}{{\bf{X}}^{\bf{T}}}{\bf{A}})} 
			\end{array} $}
	\end{equation}
		where ${{\bf{M}}_{{\cal S} \to {\cal T}}}$ is defined as
			\begin{equation}\label{eq:mstot}
			\begin{array}{c}
				(\bf M_{{{\cal S} \to {\cal T}}})_{ij} = \left\{ {\begin{array}{*{20}{l}}
						{\frac{1}{{n_s^{(c)}n_s^{(c)}}},\;\;\;{x_i},{x_j} \in {D_{\cal S}}^{(c)}}\\
						{\frac{1}{{n_t^{(r)}n_t^{(r)}}},\;\;\;{x_i},{x_j} \in {D_{\cal T}}^{(r)}}\\
						{\frac{{ - 1}}{{n_s^{(c)}n_t^{(r)}}},\;\;\;\left\{ {\begin{array}{*{20}{l}}
									{{x_i} \in {\cal D_{\cal S}}^{(c)},{x_j} \in {D_{\cal T}}^{(r)}}\\
									{{x_i} \in {\cal D_{\cal T}}^{(r)},{x_j} \in {\cal D_{\cal S}}^{(c)}}
								\end{array}} \right.}\\
							{0,\;\;\;\;\;\;\;\;\;\;\;\;otherwise}
						\end{array}} \right.
					\end{array}
				\end{equation}
Symmetrically, $Dist_{{\cal T} \to {\cal S}}^{repulsive}$ represents the sum of the distances from each target sub-domain ${D_{\cal T}}^{(c)}$ to all the the source sub-domains ${D_{\cal S}}^{(r);\;r \in \{ \{ 1...C\}  - \{ c\} \} }$ except the source sub-domain with the label $c$. Similarly, the sum of these distances is explicitly defined as:	
	\begin{equation}\label{eq:TtoS}
		\resizebox{1\hsize}{!}{%
			$Dist_{T \to S}^{repulsive} = \sum\limits_{c = 1}^C \begin{array}{l}
			{\left\| {\frac{1}{{n_s^{(c)}}}\sum\limits_{{x_i} \in {D_T}^{(c)}} {{{\bf{A}}^T}{x_i}}  - \frac{1}{{\sum\limits_{r \in \{ \{ 1...C\}  - \{ c\} \} } {n_t^{(r)}} }}\sum\limits_{{x_j} \in D_S^{(r)}} {{{\bf{A}}^T}{x_j}} } \right\|^2}\\
			= \sum\limits_{c = 1}^C {tr({{\bf{A}}^T}{\bf{X}}{{\bf{M}}_{T \to S}}{{\bf{X}}^{\bf{T}}}{\bf{A}})} 
			\end{array}  $}
	\end{equation}
	where ${{\bf{M}}_{{\cal T} \to {\cal S}}}$ is defined as
		\begin{equation}\label{eq:mstot}
			\begin{array}{c}
				(\bf M_{{{\cal T} \to {\cal S}}})_{ij} = \left\{ {\begin{array}{*{20}{l}}
						{\frac{1}{{n_t^{(c)}n_t^{(c)}}},\;\;\;{x_i},{x_j} \in {D_{\cal T}}^{(c)}}\\
						{\frac{1}{{n_s^{(r)}n_s^{(r)}}},\;\;\;{x_i},{x_j} \in {D_{\cal S}}^{(r)}}\\
						{\frac{{ - 1}}{{n_t^{(c)}n_s^{(r)}}},\;\;\;\left\{ {\begin{array}{*{20}{l}}
									{{x_i} \in {\cal D_{\cal T}}^{(c)},{x_j} \in {D_{\cal S}}^{(r)}}\\
									{{x_i} \in {\cal D_{\cal S}}^{(r)},{x_j} \in {\cal D_{\cal T}}^{(c)}}
								\end{array}} \right.}\\
							{0,\;\;\;\;\;\;\;\;\;\;\;\;otherwise}
						\end{array}} \right.
					\end{array}
				\end{equation}

Finally, we obtain
	\begin{equation}\label{eq:repulsive}
		\resizebox{0.9\hsize}{!}{%
			 ${Dist}^{repulsive} = \sum\limits_{c = 1}^C {tr({{\bf{A}}^T}{\bf{X}}({{\bf{M}}_{S \to T}} + {{\bf{M}}_{T \to S}}){{\bf{X}}^{\bf{T}}}{\bf{A}})} $}
	\end{equation}

We define ${{\bf{M}}_{\hat c}} = {{\bf{M}}_{S \to T}} + {{\bf{M}}_{T \to S}}$ as the \textit{repulsive force} constraint matrix.
While the minimization of Eq.(\ref{eq:marginal}) and Eq.(\ref{eq:conditional}) makes closer both marginal and conditional distributions between source and target, the maximization of Eq.(\ref{eq:repulsive}) increases the distances between source and target sub-domains with different labels, thereby improve the discriminative power of the underlying latent feature space.


\subsection{Label Deduction }
In a number of domain adaptation methods, \textit{e.g.},\cite{pan2008transfer,pan2011domain,long2013transfer,4967588}, the simple Nearest Neighbor (NN) classifier is applied for label deduction. In JDA, NN-based label deduction is applied twice at each iteration. NN is first applied to the target domain in order to generate the \emph{pseudo} labels of the target data and enable the computation of the conditional probability distance as defined in section 3.3. Once the optimized latent subspace NN identified, NN is then applied once again at the end of an iteration for the label prediction of the target domain. However, NN could not be a good classifier, given the fact that it is usually based on a $L2$ or $L1$ distance. It could fall short to measure the similarity of source and target domain data which may be embedded into a manifold with complex data structure. Furthermore, the cross-domain discrepancy still exists, even within a reduced latent feature space. 

To respect the underlying data manifold structure and better bridge the mismatch between the source and target domain distributions, we further propose  in this paper two consistency constraints, namely \textit{label smoothness consistency} and \textit{geometric structure consistency} for both the \emph{pseudo} and final label prediction. 



\textbf{Label Smoothness Consistency (LSC)} is defined as:

\begin{equation}\label{eq:labelconsitency}
	\begin{array}{c}
		Dis{t^{lable}} = \sum\limits_{j = 1}^C {\mathop \sum \limits_{i = 1}^{{n_s} + {n_t}} } \left\| {\bf{Y}}^{(T)}_{i,j}-{\bf{Y}}^{(0)}_{i,j} \right\|
	\end{array}
\end{equation}

where ${\bf{Y}} = {{\bf{Y}}_{\cal S}} \cup {{\bf{Y}}_{\cal T}}$,  ${\bf{Y}}_{i,j}^{(T)}$ is the probability of ${i_{th}}$ data belonging to ${j_{th}}$ class after ${T_{th}}$ iteration. ${\bf{Y}}_{i,j}^{(0)}$ is the initial prediction, and is defined as:
\vspace{-5pt}
\begin{equation}\label{eq:labelconsistency1}
\begin{array}{*{20}{l}}
{{\bf{Y}}_{{{\cal S}_{(ij)}}}^{(0)} = \left\{ {\begin{array}{*{20}{l}}
		{y_{{{\cal S}_{(ij)}}}^{(0)} = 1\;(1 \le i \le {n_s}),j = c,{y_{ij}} \in D_{\cal S}^{(c)}}\\
		{0\;\;\;\;\;\;\;\;\;\;\;else}
		\end{array}} \right.}\\
{{\bf{Y}}_{{{\cal T}_{(ij)}}}^{(0)} = \left\{ {\begin{array}{*{20}{l}}
		\begin{array}{l}
		y_{{{\cal T}_{(ij)}}}^{(0)} = 1\;(({n_s} + 1) \le i \le {n_s} + {n_t}),j = c,\\
		{y_{ij}} \in D_{\cal T}^{(c)}
		\end{array}\\
		{0\;\;\;\;\;\;\;\;\;\;\;else}
		\end{array}} \right.}
\end{array}
\end{equation}

\textbf{Geometric Structure Consistency (GSC)} is defined as: 
\begin{equation}\label{eq:YLY}
		\begin{array}{c}
		\begin{array}{l}
		\begin{array}{l}
		{{\bf{Y}}^T}{\bf{L}}{\bf{Y}} = {{\bf{Y}}^T}({\bf{I}} - {{\bf{D}}^{ - \frac{1}{2}}}{\bf{W}}{{\bf{D}}^{ - \frac{1}{2}}}){\bf{Y}} = \\
		\sum\limits_{i = 1}^{{n_s} + {n_t}} {{d_{ii}}{{\left( {\frac{{{y_i}}}{{\sqrt {{{\bf{d}}_{ii}}} }}} \right)}^2}}  - \sum\limits_{i,j = 1}^{{n_s} + {n_t}} {{{\bf{d}}_{ii}}{{\left( {\frac{{{y_i}}}{{\sqrt {{{\bf{d}}_i}} }}\frac{{{y_j}}}{{\sqrt {{{\bf{d}}_j}} }}} \right)}^2}} {{\bf{w}}_{ij}}\;\\
		= \frac{1}{2}\sum\limits_{i,j = 1}^{{n_s} + {n_t}} {{{\bf{w}}_{ij}}{{\left( {\frac{{{y_i}}}{{\sqrt {{{\bf{d}}_{ii}}} }} - \frac{{{y_j}}}{{\sqrt {{{\bf{d}}_{jj}}} }}} \right)}^2}} 
		\end{array}
		\end{array},
		\end{array}
		\end{equation}

where  ${\bf{W}} = {[{w_{ij}}]_{({n_s} + {n_t}) \times ({n_s} + {n_t})}}$ is an affinity matrix \cite{NIPS2001_2092}, with   ${w_{ij}}$ giving the affinity between two samples $i$ and $j$ and  defined as ${w_{ij}} = \exp ( - \frac{{{{\left\| {{x_i} - {x_j}} \right\|}^2}}}{{2{\sigma ^2}}})$ if $i \ne j$ and ${w_{ii}} = 0$ otherwise, ${\bf{D}} = diag\{ {d_{11}}...{d_{({n_s} + {n_t}),({n_s} + {n_t})}}\} $ is the degree matrix with ${d_{ii}} = \sum\nolimits_j {{w_{ij}}} $. When Eq.(\ref{eq:YLY}) is minimized, the geometric structure consistency ensures that the label space does not change too much between nearby data.

\subsection{Learning Algorithm}
Our proposed domain adaptation integrates the marginal and conditional distribution and repulsive force, as well as the final label prediction using both label smoothness and geometric structure consistencies. Our model is defined as:
\begin{equation}\label{eq:ours_physical}
		\resizebox{0.96\hsize}{!}{%
	$\begin{array}{*{20}{l}}
		{\min (Dis{t^{marginal}} + Dis{t^{conditional}} + Dis{t^{label}} + {{\bf{Y}}^T}L{\bf{Y}})}\\ + 
		\max (Dist^{repulsive})
	\end{array}$}
\end{equation}

It can be re-written mathematically as:
\begin{equation}\label{eq:ours_math}
	\resizebox{1\hsize}{!}{%
$\begin{array}{*{20}{l}}
{\begin{array}{*{20}{l}}
	{\mathop {\min }\limits_{{{\bf{A}}^T}{\bf{XH}}{{\bf{X}}^T}{\bf{A}} = {\bf{I}}} \left( {\begin{array}{*{20}{l}}
			{\sum\limits_{c = 0}^C {tr({{\bf{A}}^T}{\bf{X}}{{\bf{M}}_c}{{\bf{X}}^T}A)}  + \lambda \left\| {\bf{A}} \right\|_F^2}\\
			{ + \sum\limits_{j = 1}^C {\sum\limits_{i = 1}^{{n_s} + {n_t}} {\left\| {{\bf{Y}}_{ij}^{(T)} - {\bf{Y}}_{ij}^{(0)}} \right\|} }  + {{\bf{Y}}^T}{\bf{LY}}}
			\end{array}} \right)}\\
	{ + \mathop {\max }\limits_{{{\bf{A}}^T}{\bf{XH}}{{\bf{X}}^T}{\bf{A}} = {\bf{I}}} tr({{\bf{A}}^T}{\bf{X}}{{\bf{M}}_{{\bf{\hat c}}}}{{\bf{X}}^T}{\bf{A}})}
	\end{array}}
\end{array}$}
\end{equation}

Direct solution to this problem is nontrivial. We divide it into two sub-problems:  (1) $ \mathop {\min }\limits_{{{\bf{A}}^T}{\bf{XH}}{{\bf{X}}^T}{\bf{A}} = {\bf{I}}} \left( {\sum\limits_{c = 0}^C {tr({{\bf{A}}^T}{\bf{X}}{{\bf{M}}_{cyd}}{{\bf{X}}^T}A)}  + \lambda \left\| {\bf{A}} \right\|_F^2} \right) $, where ${{\bf{M}}_{cyd}} = \sum\limits_{c = 0}^C {{{\bf{M}}_c} - {{\bf{M}}_{{\bf{\hat c}}}}} $ and (2)  $\mathop {\min }\limits_{{{\bf{A}}^T}{\bf{XH}}{{\bf{X}}^T}{\bf{A}} = {\bf{I}}} \left( {\sum\limits_{j = 1}^C {\sum\limits_{i = 1}^{{n_s} + {n_t}} {\left\| {{\bf{Y}}_{ij}^{(T)} - {\bf{Y}}_{ij}^{(0)}} \right\|} }  + {{\bf{Y}}^T}{\bf{L}}{\bf{Y}}} \right)$. These two sub-problems are then iteratively optimized. 

The first sub-problem, as explained in  JDA, amounts to solving the generalized eigendecomposition problem,\textit{i.e.},  $({\bf{X}}{\bf{M_{cyd}}}{{\bf{X}}^T} + \lambda {\bf{I}}){\bf{A}} = {\bf{XH}}{{\bf{X}}^T}{\bf{A}}\Phi $. Then, we obtain the adaptation matrix $\boldsymbol{A}$ and the underlying embedding space $\boldsymbol{Z}$. 

The  second sub-problem is  nontrivial. Inspired by the solution proposed in   \cite{Zhou04learningwith} \cite{6341755} \cite{6619251}, the minimum is approached where the derivative of the function is zero.  An approximate solution can be provided by: 
\vspace{-5pt}
\begin{equation}\label{eq:Y_optimal}
	{{\bf{Y}}^ \star } = {({\bf{D}} - \alpha {\bf{W}})^{ - 1}}{Y^{(0)}}
\end{equation}
where  $Y^\star$ is the probability of prediction of the target domain corresponding to different class labels. 


The complete learning algorithm  is summarized in Algorithm 1.

\begin{algorithm}[!h]
	\caption{Close yet Discriminative Domain Adaptation (CDDA)}
	\KwIn{Data $\bf{X}$, Source domain label ${\bf{Y}}_{\cal S}$, subspace bases $k$, iterations $T$, regularization parameter $\lambda $ and $\alpha $}
	\While{$\sim isempty(\bf{X},{{\bf{Y}}_{\cal S}})$ and $t<T$	}{
			\textbf{Step {1}}:  Construct ${\bf{M}}_c$ and  ${\bf M}_{\hat{c}}$ ;\\
			\textbf{Step 2}: Projection space calculation \\
			
			(i) Calculate ${{\bf{M}}_{cyd}} = {{\bf{M}}_c} -{\bf M}_{\hat{c}} $;\\
			(ii) Solve the generalized eigendecomposition problem as in Eq.(\ref{eq:ours_math}) and obtain adaptation matrix $\bf A$, then  embed data via the transformation, $\bf{Z} = {{\bf{A}}^T}{\bf{X}}$\;
			
			\textbf{Step 3}: Labels deduction \\
			\eIf{ $ \sim isempty(\bf{Z},{{\bf{Y}}_{\cal S}})$ }{
				(i) construct the label matrix ${{\bf{Y}}^{(0)}}$\;
				(ii) initialize the graph G, construct the affinity matrix $\boldsymbol{W}$ and diagonal matrix $\boldsymbol{D}$\;
				(iii) obtain ${{\rm{{\bf{Y}}}}_{final}}$ in solving Eq.(\ref{eq:Y_optimal})\;
			}{
			\textbf{break}\;
		}
		
		\textbf{Step  4}: update pseudo target labels $\{ {\bf{Y}}_{\cal T}^{(T)} = {{\bf{Y}}_{final}}\left[ {:,({n_s} + 1):({n_s} + {n_t})} \right]\} $;\\
		\textbf{Step  5}: Return to Step1;  $t=t+1$;\\               
	
}
\KwOut{Adaptation matrix ${\bf{A}}$, embedding ${\bf{Z}}$, Target domain labels ${\bf{Y}}_{\cal T}^{(T)}$}
\end{algorithm}

\section{Experiments}
In this section, we validate the effectiveness of our proposed  domain adaptation model, \textit{i.e.}, CDDA, on several datasets for cross-domain image classification task.
\subsection{Benchmarks}
In domain adaptation, USPS+MINIST, COIL20, PIE and office+Caltech are standard benchmarks for the purpose of evaluation and comparison with state of the art. In this paper, we follow the data preparation as most previous works. We construct 36 datasets for different image classification tasks. They are: (1) the \textbf{USPS} and \textbf{MINIST} datasets of digits, but with different distribution probabilities. We built the cross-domains as: \emph{USPS vs MNIST} and \emph{MNIST vs USPS}; (2) the \textbf{COIL20} dataset with 20 classes, split into \emph{ COIL1 vs COIL2} and \emph{COIL2 vs COIL1}; (3) the  \textbf{PIE} face database with different face {poses}, of which five subsets are selected, denoted as PIE1, PIE2, \textit{etc}., resulting in $5 \times 4=20$ domain adaptation tasks, \textit{i.e.}, \emph{PIE1 vs PIE 2} $\dots$ \emph{PIE5 vs PIE 4}; (4) \textbf{Office} and \textbf{Caltech-256}. Office contains three real-world datasets: \textbf{Amazon}(images downloaded from online merchants), \textbf{Webcam}(low resolution images) and \textbf{DSLR}( high-resolution images by digital web camera). \textbf{Caltech-256} is standard dataset for object recognition, which contains 30,607 images for 31 categories. We denote the dataset \textbf{Amazon},\textbf{Webcam},\textbf{DSLR},and \textbf{Caltech-256} as \textbf{A},\textbf{W},\textbf{D},and \textbf{C}, respectively.  $4\times 3=12$ domain adaptation tasks can then be constructed, namely \emph{A} $\rightarrow$ \emph{W} $\dots$ \emph{C} $\rightarrow$ \emph{D}, respectively.

\subsection{Baseline Methods}
The proposed CDDA method is compared with six methods of the literature, excluding only CNN-based works, given the fact that we are not using deep features. They are: (1)1-Nearest Neighbor Classifier(NN); (2) Principal Component Analysis (PCA) +NN; (3) Geodesic Flow Kernel(GFK) \cite{gong2012geodesic} + NN; (4) Transfer Component Analysis(TCA) \cite{pan2011domain} +NN; (5)Transfer Subspace Learning(TSL) \cite{4967588} +NN; (6) Joint Domain Adaptation (JDA) \cite{long2013transfer} +NN. Note that TCA and TSL can be viewed as special case of JDA with $C=0$, and JDA a special case of the proposed CDDA method when the \textit{repulsive force} domain adaptation is ignored and the label generation is simply based on NN instead of the label propagation with label smoothness and geometric structure consistency constraints.  

All the reported performance scores of the six methods of the literature are directly collected from the authors' publication. They are assumed to be their \emph{best} performance.
%
%
%

\subsection{Experimental Setup}
For the problem of domain adaptation, it is not possible to tune a set of optimal hyper-parameters, given the fact that the target domain has no labeled data. Following the setting of JDA, we also evaluate the proposed CDDA by empirically searching the parameter space for the \emph{optimal} settings. Specifically, the proposed CDDA method has three hyper-parameters, \textit{i.e.}, the subspace dimension $k$, regularization parameters $\lambda $ and $\alpha $. In  our experiments, we set $k = 100$ and 1) $\lambda  = 0.1$, and $\alpha  = 0.99$ for \textbf{USPS}, \textbf{MNIST} and \textbf{COIL20} , 2) $\lambda  = 0.1$, $\alpha  = 0.2$ for \textbf{PIE}, 3) $\lambda  = 1$, $\alpha  = 0.99$ for \textbf{Office} and \textbf{Caltech-256}.

In our experiment, {\emph{accuracy}}  on the test dataset is the evaluation measurement. It is widely used in literature, \textit{e.g.},\cite{pan2008transfer,long2013transfer,long2015learning}, \textit{etc}.

\begin{equation}\label{eq:accuracy}
	\begin{array}{c}
		Accuracy = \frac{{\left| {x:x \in {D_T} \wedge \hat y(x) = y(x)} \right|}}{{\left| {x:x \in {D_T}} \right|}}
	\end{array}
\end{equation}
where ${\cal{D_T}}$ is the target domain treated as test data, ${\hat{y}(x)}$ is the predicted label and ${y(x)}$ is the ground truth label for a test data  $x$.

\subsection{Experimental Results and Discussion}
The classification accuracies of the proposed CDDA method and the six baseline methods  are shown in Table.1. and illustrated  in Fig.1. for the clarity of comparison. 
\begin{figure*}[h!]
	\centering
	\label{fig:acc}
	\includegraphics[width=1\linewidth]{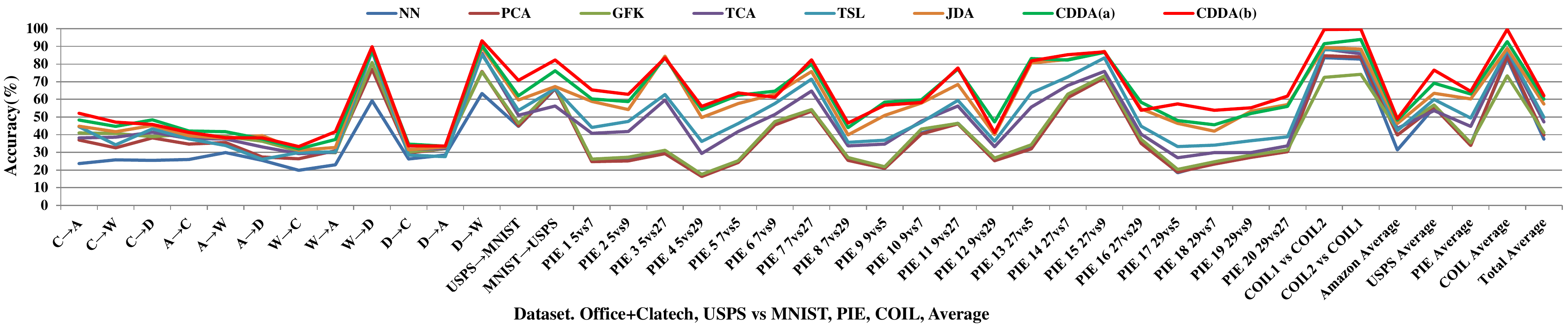}
	\vspace{-7pt}
	\caption { Accuracy($\% $) on the 36 cross-domain image classification tasks using 4 different image datasets, each under different difficulty for knowledge transfer. } 
\end{figure*} 
\begin{table}[h!]
	\centering
	\label{tab:acc}
	\caption{Quantitative comparisons with the baseline methods: Accuracy($\% $) on 36 cross-domain image classifications on four different datasets}
	\vspace{1pt}
	\resizebox{\columnwidth}{!}{%
		\begin{tabular}{|l |c |c |c| c |c |c | c| c|}\hline
			\toprule
			
			Datasets& {NN} & {PCA} & {GFK} & {TCA} & {TSL}  & {JDA} & \textbf{CDDA(a)} & \textbf{CDDA(b)} \\
			\midrule
			
			USPS \emph{vs} MNIST& 44.70&	44.95&	46.45&	51.05& 53.75&	59.65&	62.05&	\textbf{ 70.75}
			\\
			MNIST \emph{vs} USPS& 65.94&	66.22&	67.22&	56.28&	66.06&	67.28&	76.22&	\textbf{ 82.33}
			\\
			\hline
			COIL1 \emph{vs} COIL2& 83.61&	84.72&	72.50&	88.47&	88.06&	89.31&	91.53&	\textbf{ 99.58}
			\\
			COIL2 \emph{vs} COIL1& 82.78&	84.03&	74.17&	85.83&	87.92&	88.47&	93.89&	\textbf{ 99.72}
			\\
			\hline
			
			\vspace{1pt}  
			
			PIE1 \emph{vs} PIE2&  26.09&	24.80&	26.15&	40.76&	44.08&	58.81&	60.22&	\textbf{ 65.32}
			\\
			PIE1 \emph{vs} PIE3&  26.59&	25.18&	27.27&	41.79&	47.49&	54.23&	58.70&  \textbf{   62.81}
			\\
			PIE1 \emph{vs} PIE4 & 30.67&29.26&	31.15&	59.63&	62.78&\textbf{84.50}& 83.48& 83.54
			\\
			PIE1 \emph{vs} PIE5&   16.67&	16.30&	17.59&	29.35&	36.15&	49.75&	54.17&	\textbf{ 56.07}
			\\
			PIE2 \emph{vs} PIE1&   24.49&	24.22&	25.24&	41.81&	46.28&	57.62&	62.33&	\textbf{ 63.69}
			\\
			PIE2 \emph{vs} PIE3&   46.63&	45.53&	47.37&	51.47&	57.60&	62.93&	\textbf{ 64.64}&	61.27
			\\
			PIE2 \emph{vs} PIE4&   54.07&	53.35&	54.25&	64.73&	71.43&	75.82&	79.90	&\textbf{ 82.37}
			\\
			PIE2 \emph{vs} PIE5&   26.53&	25.43&	27.08&	33.70&	35.66&	39.89&	44.00&	\textbf{ 46.63}
			\\
			PIE3 \emph{vs} PIE1&   21.37&	20.95&	21.82&	34.69&	36.94&	50.96&	\textbf{ 58.46}&	56.72
			\\
			PIE3 \emph{vs} PIE2&   41.01&	40.45&	43.16&	47.70&	47.02&	57.95&	\textbf{ 59.73}&	58.26
			\\
			PIE3 \emph{vs} PIE4&   46.53&	46.14&	46.41&	56.23&	59.45&	68.45&	77.20&	\textbf{ 77.83}
			\\
			PIE3 \emph{vs} PIE5&   26.23&	25.31&	26.78&	33.15&	36.34&	39.95&	\textbf{ 47.24}&	41.24
			\\
			PIE4 \emph{vs} PIE1&   32.95&	31.96&	34.24&	55.64&	63.66&	80.58&	\textbf{ 83.10}&	81.84
			\\
			PIE4 \emph{vs} PIE2&  62.68&	60.96&	62.92&	67.83&	72.68&	82.63&	82.26&	\textbf{ 85.27}
			\\
			PIE4 \emph{vs} PIE3&   73.22&	72.18&	73.35&	75.86&	83.52&	\textbf{87.25}&	86.64&	86.95
			\\
			PIE4 \emph{vs} PIE5&   37.19&	35.11&	37.38&	40.26&	44.79&	54.66&	\textbf{ 58.33}&	53.80
			\\
			PIE5 \emph{vs} PIE1&  18.49&	18.85&	20.35&	26.98&	33.28&	46.46&	48.02&	\textbf{ 57.44}
			\\
			PIE5 \emph{vs} PIE2&   24.19&	23.39&	24.62&	29.90&	34.13&	42.05&	45.61&	\textbf{ 53.84}
			\\
			PIE5 \emph{vs} PIE3&  28.31&	27.21&	28.49&	29.9&	36.58&	53.31&	52.02&	\textbf{ 55.27}
			\\
			PIE5 \emph{vs} PIE4&   31.24&	30.34&	31.33&	33.64&	38.75&	57.01&	55.99&	\textbf{ 61.82}
			\\
			\hline
			
			C $\rightarrow$ A&  23.70&	36.95&	41.02&	38.20&	44.47&	44.78&	48.33&	\textbf{ 52.09}
			\\
			C $\rightarrow$ W&  25.76&	32.54&	40.68&	38.64&	34.24&	41.69&	44.75&	\textbf{ 47.12}
			\\
			C $\rightarrow$ D&  25.48&	38.22&	38.85&	41.40&	43.31&	45.22&	48.41&	\textbf{ 45.86}
			\\
			A $\rightarrow$ C&  26.00&	34.73&	40.25&	37.76&	37.58&	39.36&	42.12&	\textbf{ 41.32}
			\\
			A $\rightarrow$ W&  29.83&	35.59&	38.98&	37.63&	33.90&	37.97&	\textbf{41.69}&	38.31
			\\
			A $\rightarrow$ D&  25.48&	27.39&	36.31&	33.12&	26.11&	\textbf{39.49}&	37.58&	38.22
			\\
			W $\rightarrow$ C&  19.86&	26.36&	30.72&	29.30&	29.83&	31.17&	31.97&	\textbf{ 33.30}
			\\
			W $\rightarrow$ A&  22.96&	31.00&	29.75&	30.06&	30.27&	32.78&	37.27&	\textbf{ 41.75}
			\\
			W $\rightarrow$ D&  59.24&	77.07&	80.89&	87.26&	87.26&	89.17&	87.90&	\textbf{ 89.81}
			\\
			D $\rightarrow$ C&  26.27&	29.65&	30.28&	31.70&	28.50&	31.52&	\textbf{ 34.64}&	33.66
			\\
			D $\rightarrow$ A & 28.50&	32.05&	32.05&	32.15&	27.56&	33.09&	33.51&	\textbf{ 33.61}
			\\
			D $\rightarrow$ W & 63.39&	75.93&	75.59&	86.10&	85.42&	89.49&	90.51&	\textbf{ 93.22}
			\\
			\hline
			\hline
			Average (USPS)& 55.32&	55.59&	56.84&	53.67&	59.90&	63.47&	69.14&	\textbf{ 76.54}
			\\
			Average (COIL)& 83.20&	84.38&	73.34&	87.15&	87.99&	88.89&	92.71&	\textbf{ 99.65}
			\\
			Average (PIE)& 34.76&	33.85&	35.35&	44.75&	49.43&	60.24&	63.10&	\textbf{ 64.60}
			\\
			Average (Amazon)&  31.37&	39.79&	42.95&	43.61&	42.37&	46.31&	48.22&	\textbf{ 49.02}
			\\
			
			Overall Average&     37.46&	39.84&	41.19&	47.22&	49.80&	57.37&	60.12&	\textbf{ 62.02}
			\\
			\bottomrule
		\end{tabular}}
		
	\end{table}
	
	\begin{figure*}[h!]
		\label{fig:analysis}
		\begin{center}
			\begin{tabular}{c}
				\includegraphics[width=1\linewidth]{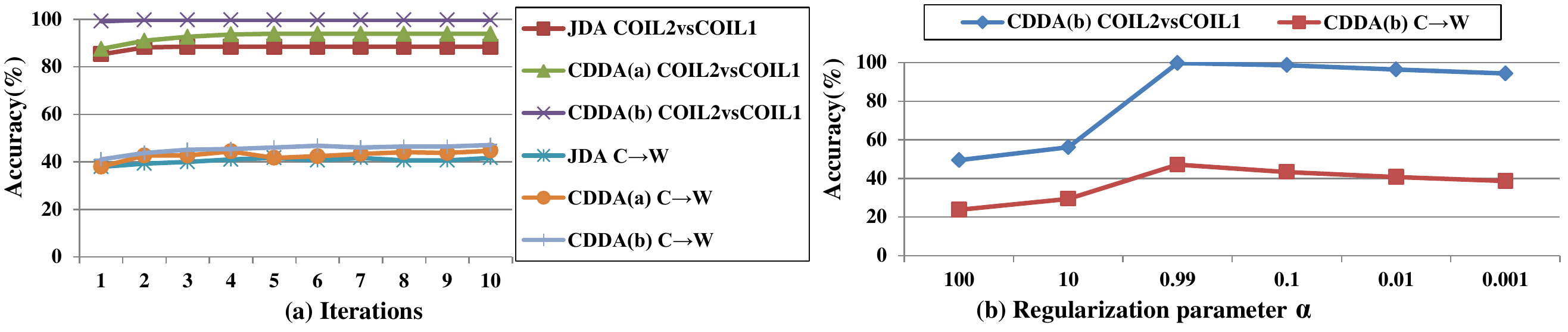}
			\end{tabular}
		\end{center}
		\vspace{-11pt}
		\caption { Parameter sensitivity and convergence analysis: (a) accuracy w.r.t $\#$iterations; (b) accuracy w.r.t regularization parameter $\alpha$. } 
	\end{figure*} 
	In Table.1, the highest accuracy for each cross-domain adaptation task is highlighted in  bold. For a better understanding of the proposed CDDA,  we evaluate the proposed CDDA method using two settings: (1) \textbf{CDDA(a)} where simple NN is used as label predictor instead of the proposed label propagation; and (2)  \textbf{CDDA(b)} where the proposed label propagation is activated for the prediction of target data labels. As CDDA is reduced to JDA when repulsive force domain adaptation and label propagation are not integrated, the setting CDDA(a) enables to quantify the contribution of adding the repulsive force domain adaptation w.r.t. JDA whereas the setting CDDA(b) makes it possible to evidence the contribution of the proposed label propagation in comparison with CDDA(a) and highlight the overall behavior of the proposed method.
    
    As can be seen  in Table.1 , the proposed CDDA depicts an overall average accuracy of  $\bf 60.12\%$ and $\bf 62.92\%$, respectively,  with respect to the above two settings. They both outperform the six baseline algorithms with a large margin. With the repulsive force integrated and NN as label predictor, CDDA(a) outperforms JDA on 30 cross-domain tasks out of 36 and improves JDA's overall average accuracy by roughly 3 points, thereby demonstrating the effectiveness of the proposed repulsive force domain adaptation. Now, in adopting the proposed label propagation under the constraint of both label smoothness and geometric structure consistency, CDDA(b) further improves CDDA(a) by roughly 2 points in terms of overall average accuracy and outperforms JDA by more than 4 points. 
    Compared with the baseline methods, the proposed CDDA method consistently shows its superiority  and depicts the best average  accuracy over all the four datasets (USPS+MINIST, COIL20, PIE, Amazon).  As can be seen in Fig.2,  \textbf{CDDA(b)} as represented by the red curve is on the top of the other curves along  the axis of 36 cross-domain image classification tasks.  It is worth noting  that the proposed  CDDA depicts $\bf 99.65$ accuracy on \textbf{COIL20}; This is rather an unexpected impressive score given the unsupervised nature of the domain adaptation for the target domain.

Using  \emph{ COIL2 vs COIL1}, and \emph{C} $\rightarrow$ \emph{W} datasets, we also empirically check the convergence  and the sensitivity of the proposed CDDA  with respect to the hyper-parameters.  Similar trends can be observed on all the other datasets.
	
The accuracy w.r.t. $\#$iterations is shown in Fig.3 (a). As can be seen there, the performance of the proposed CDDA  along with JDA  becomes stable after about 10 iterations. 
	
In the experiment, CDDA have two settings: two parameters ($k$ and $\lambda$) in \textbf{CDDA(a)} and three ($k$, $\lambda$ and $\alpha$) in  \textbf{CDDA(b)}. The accuracy variation w.r.t regularization parameter $\alpha$ is shown in Fig.3 (b), which indicates \textbf{ CDDA(b)} achieves the best performance when $\alpha$ is close to 0.99 in COIL20 and the performance is more or less stable  when  $\alpha$ is less than 0.99. Given a novel dataset, we tune the parameter $\alpha$ in the range  [0.001,1]. 
	For instance, in the \textbf{PIE} database, we set the optimal $\alpha$  to 0.2. The other parameters, \textit{i.e.},  $k$ and $\lambda$, also converge.  Their behavior is not shown here  due to space limitation.

\section{Conclusion and Future Work}
In this paper, we have proposed a Close yet Discriminative Domain Adaptation (CDDA) method based on feature representation. Comprehensive experiments on 36 cross-domain datasets highlight the interest of reinforcing the data discriminative properties within the model and label propagation in respect of the geometric structure of the underlying data manifold, and  verify the effectiveness  of proposed method compared with six baseline methods of the literature.  


Our future work will concentrate on embedding the proposed method in  deep  networks and study other vision tasks, \textit{e.g.}, object detection, within the setting of transfer learning.

{\small
\bibliographystyle{hacm}
\bibliography{egbib}
}
\end{CJK*}

\end{document}